\documentclass{article} %
\usepackage{iclr2027_conference,times}

\usepackage{amsmath,amsfonts,bm}

\def\eqref#1{equation~\ref{#1}}

\def\1{\bm{1}}

\DeclareMathAlphabet{\mathsfit}{\encodingdefault}{\sfdefault}{m}{sl}
\SetMathAlphabet{\mathsfit}{bold}{\encodingdefault}{\sfdefault}{bx}{n}

\usepackage{hyperref}
\usepackage{xurl}
\usepackage{cleveref}
\usepackage{url}
\usepackage{booktabs}
\usepackage{tabularx}
\usepackage{graphicx}
\usepackage{wrapfig}
\usepackage{enumitem}

\usepackage{fontawesome5}

\usepackage[most]{tcolorbox}
\tcbuselibrary{listings, breakable, skins}

\newtcblisting{promptbox}[1]{
  listing only,
  breakable,
  enhanced,
  colback=gray!4,
  colframe=gray!55!black,
  coltitle=white,
  colbacktitle=gray!55!black,
  fonttitle=\bfseries\small,
  title=#1,
  boxrule=0.4pt,
  arc=1pt,
  left=4pt, right=4pt, top=3pt, bottom=3pt,
  listing options={
    basicstyle=\ttfamily\scriptsize,
    breaklines=true,
    breakindent=0pt,
    columns=fullflexible,
    keepspaces=true,
    showstringspaces=false,
    xleftmargin=0pt,
    escapeinside={(*}{*)},
  },
}

\newcommand{\Sref}[1]{\S\ref{#1}}
\newcommand{\sref}[1]{\S\ref{#1}}

\newcommand{\inst}[1]{\ensuremath{^{#1}}}
\newcommand{\authorentry}[2]{\mbox{#1\inst{#2}}}

\title{InterviewPlayground: A Validated \\Simulation Environment for Evaluating \\AI Interviewers}

\author{%
  \parbox[t]{0.95\textwidth}{\centering
    \authorentry{Jonathan Ivey}{1},~~\allowbreak\ %
    \authorentry{Aimee Liang}{1},~~\allowbreak\ %
    \authorentry{Arthur Y.S. Wang}{2},~~\allowbreak\ %
    \authorentry{Madeline Mandell}{2},~~\allowbreak\ %
    \authorentry{Ziang Xiao}{1},~~\allowbreak\ %
    \authorentry{Anjalie Field}{1}
    \\[0.5em]
    {\normalfont\small
      \inst{1}Johns Hopkins University \quad
      \inst{2}Listen Labs
    }\\[0.2em]
    {\normalfont\small\href{https://github.com/jonathanivey/interviewplayground}{\faGithub\ \texttt{github.com/jonathanivey/interviewplayground}}}
  }%
}

\iclrfinalcopy %
\begin{document}

\maketitle

\begin{abstract}

Increasingly, AI interviewers are being developed to elicit open-ended responses in applications like market research, public polling, preference elicitation, and social science research. However, evaluating AI interviewers is challenging because they function in extended, multi-turn interactions where they must adapt to participant behaviors.
To address this need, we develop \textit{InterviewPlayground}, a simulation environment for evaluating AI interviewers using simulated study participants whose behaviors are grounded in social theory. Simulated studies in InterviewPlayground produce an \textit{InterviewReportCard}, which assesses the performance of AI interviewers using a suite of validated measures. To test whether our simulation-based evaluations predict performance with human participants, we conduct 15 real qualitative studies with five AI interviewers, three interview topics, and 450 human participants and compare them to simulated studies in InterviewPlayground. We find that AI interviewer performance in InterviewPlayground predicts performance in human studies with an average Pearson correlation of 0.86 across 12 measures, and the simulated interactions from InterviewPlayground reproduce key findings from behavioral analysis of AI interviewers in the human studies. Together, these findings support the validity of InterviewPlayground in assessing AI interviewer performance and examining potential failure modes.  %
Our work contributes a simulation environment for AI interviewers supported with empirical validation, and more broadly, a roadmap for future work to develop validated, simulation-based evaluations of conversational AI systems.

\end{abstract}
\begin{figure}[h]
    \centering
    \includegraphics[width=1\linewidth]{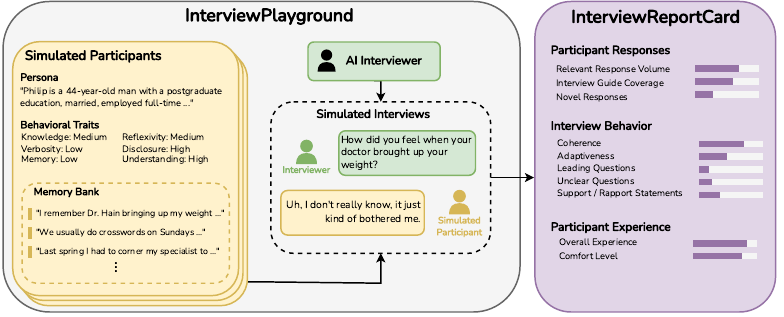}
    
    \caption{InterviewPlayground is a simulation environment for evaluating AI interviewers with simulated studies. In the simulated studies, AI interviewers aim to answer a research question by interacting with simulated participants. The simulated study produces an InterviewReportCard, which assesses the participant responses, interviewer behavior, and participant experience.}
    \label{fig:teaser}
\end{figure}

\section{Introduction}

Interviews provide rich insights into people's experiences, motivations, and behaviors through extended dialogues where an interviewer carefully constructs questions and adapts to participant responses. They are widely used in applications like market research \citep{chopra_conducting_2023}, preference elicitation \citep{choudhury_bed-llm_2026}, public polling \citep{jiang_communitybots_2023}, and social science research \citep{cuevas_collecting_2025, liu_mimitalk_2025}. Increasingly, conversational AI systems are being used in these interview settings to reduce cost and increase scale \citep{xiao_tell_2020, huang_what_2026, park_llm_2026}, but evaluating these dynamic, multi-turn systems remains challenging.

Some studies evaluate AI interviewers with human participants, but they are expensive and time-consuming to recruit, leading many to only evaluate with small, single-domain pilot studies \citep{xiao_tell_2020, jiang_communitybots_2023, hu_designing_2024, li_lm-interview_2024, cuevas_collecting_2025, wuttke_ai_2025, liu_mimitalk_2025, gardhus_ainterviewer_2026, wen_interflow_2026}, reducing generalizability and making it difficult to compare systems. Conversational AI in other domains like customer service \citep{yao_-bench_2024}, software engineering \citep{xu_theagentcompany_2025}, and clinical NLP \citep{kyung_patientsim_2025, schmidgall_agentclinic_2025} face similar challenges that researchers have attempted to address with simulation-based evaluations. However, these simulations often fail to satisfy one or more desiderata outlined by \citet{liao2025rethinkingmodelevaluationnarrowing}: (1) simulated user behaviors and target outcomes should be informed by research, (2) simulation-based evaluations should be validated with real user data, and (3) authors should specify suitable situations to use the simulations. Most commonly, these simulations lack validation with real users, leading to a Sim2Real gap \citep{zhou_mind_2026, liu2026syntheticusersrealdifferences}.

In this work, we develop a simulation environment for AI interviewers through a five-step process designed to address these desiderata. First, we review existing research on interview participants to identify participant behaviors that are most influential on an interview's process and outcomes. Second, we use this review to develop and operationalize a cognitive model of interview participants grounded in social theory. Third, we use this cognitive model to create InterviewPlayground, an environment with three simulation settings based on real studies in obesity and weight management, Asian American identity and politics, and generative AI in knowledge work. Fourth, we create and validate InterviewReportCard, an evaluation suite that measures three key dimensions of an interview: participant responses, interviewer behavior, and participant experience. Finally, we collect evidence supporting InterviewPlayground's predictive validity by conducting 15 real qualitative studies with five AI interviewers, the three interview topics from our simulation settings, and 450 human participants and comparing them to simulated studies in InterviewPlayground.

We find that InterviewPlayground's simulated studies \textbf{predict quantitative evaluations of AI interviewers in human studies with an average Pearson correlation of 0.86 across 12 measures}. They also reproduce key findings from behavioral analysis of AI interviewers in human studies, allowing researchers to identify potential failure modes and interaction patterns. Additionally, we find the quantitative results are robust to the choice of simulator model, including models with as few as 4B parameters, and expert judges rate simulated interview transcripts as realistic as transcripts from human participants. Together, these results support the use of InterviewPlayground's three simulation settings for testing and evaluating AI interviewers. They also demonstrate how InterviewPlayground could be used to create and validate additional simulation settings with new interview topics and participant populations. More broadly, our five-step process provides a roadmap for future work to develop validated, simulation-based evaluations for other conversational AI systems.

\section{Related Work}
\paragraph{Evaluating AI Interviewers} Interviews are extended, multi-turn interactions where interviewers must carefully construct questions and adapt to participant responses. The complexity of this interaction makes evaluating AI interviewers inherently challenging. Prior work primarily addresses this problem by evaluating with human participants, which is expensive and time-consuming. As a result, most evaluations are limited to single-domain studies of a specific AI design \citep{xiao_tell_2020, jiang_communitybots_2023, hu_designing_2024, li_lm-interview_2024, cuevas_collecting_2025, wuttke_ai_2025, liu_mimitalk_2025, gardhus_ainterviewer_2026, wen_interflow_2026}, which limits generalizability and makes it difficult to compare multiple systems. Some prior work has addressed these challenges with basic simulations for preliminary evaluation, but their simulations are not grounded in existing research on interview participants, simulate all participants with the same ``professional'' and ``confident'' personas, and lack evidence that their evaluations correlate with the performance of AI interviewers in human studies \citep{anugraha_sparkme_2026}. Works in other domains like customer service \citep{yao_-bench_2024}, software engineering \citep{xu_theagentcompany_2025}, and clinical NLP \citep{kyung_patientsim_2025, schmidgall_agentclinic_2025} have developed user simulations specific to their tasks, but they similarly lack validation that performance in their simulations correlates with real-world performance, leading to a Sim2Real gap \citep{zhou_mind_2026, liu2026syntheticusersrealdifferences}. 
In this work, we develop InterviewPlayground, a simulation environment for evaluating AI interviewers with simulated study participants whose behaviors are grounded in social theory. We collect evidence supporting InterviewPlayground's validity by conducting 15 real qualitative studies with five AI interviewers, three interview topics, and 450 human participants and comparing them to simulated studies in InterviewPlayground.

\paragraph{Simulating Qualitative Research Participants}

Prior work on simulating interview participants has primarily focused on predicting the information that a human participant would produce in a qualitative interview \citep{hamalainen_evaluating_2023, kapania_simulacrum_2025}. Accordingly, their simulation designs focus on predicting what information a participant can provide. By contrast, our simulation environment aims to evaluate interviewers, and its design is focused on the factors in the literature that most strongly influence the process and outcomes of an interview. Rather than try to predict the information that a participant can provide, we instantiate simulated participants with existing information from prior studies and evaluate an interviewer's ability to elicit it. 

\section{InterviewPlayground}

Given an AI interviewer, we aim to construct an environment that assesses how it will perform in real-world settings where it must answer a set of research questions by interacting with a sample of human participants. In these settings, human participants will vary both in their behavior and the insights that they can offer. Thus, we construct an environment in which simulated participants mimic this behavior.
To ground this environment in prior research and social theory, we first review literature on qualitative interviews to identify the participant behaviors most influential to interviewer processes and outcomes (\Sref{sec:theoretical-grounding}). Then, based on that review, we develop a cognitive model of interview participants that simulates those behaviors (\Sref{sec:cognitive-model}) and a simulation environment where AI interviewers can interact with them (\Sref{sec:interview-environment}). Finally, we use our environment to create three simulation settings based on published studies in obesity and weight management, Asian American identity and politics, and generative AI in knowledge work (\Sref{sec:study-presets}).

\subsection{Theoretical Grounding}
\label{sec:theoretical-grounding}

To understand what participant behaviors we should simulate, we first review relevant social theory and existing literature on qualitative interviews. We identify descriptions of participant behaviors that directly influence an interviewer's actions or the quality of responses. Across our review, there is broad agreement on the general behavior of an interview participant: they receive a question, attempt to recall relevant information, and produce a response \citep{kvale_interviews_2009}. However, we also identify 20 specific participant characteristics that directly influence these behaviors. We then iteratively refine this list, merging closely related characteristics into six broad behavioral traits: 

\begin{itemize}[nosep, leftmargin=*]
    \item \textbf{Knowledge}: the participant's experience or familiarity with the topic of interest.
    \item \textbf{Understanding}: a participant's ability to understand and interpret the interviewer's questions and the purpose of the interview.
    \item \textbf{Reflexivity}: the extent to which a participant is able to self-reflect and self-analyze.
    \item \textbf{Memory}: a participant's ability to accurately recall information and experiences.
    \item \textbf{Verbosity}: the length of a participant's answers.
    \item \textbf{Disclosure}: the extent to which a participant is willing to provide potentially sensitive information.
\end{itemize}

This process also produces two other traits: hostile non-cooperation and mental and physical impairment. However, due to their complexity, high stakes, and relative rarity, we leave the simulation of these traits to future work. Additional details of the review are included in \Sref{app:lit-review}.

\subsection{Cognitive Model}
\label{sec:cognitive-model}

\begin{wrapfigure}{r}{0.45\textwidth}
  \centering
  \vspace{-10pt} %
  \includegraphics[width=\linewidth]{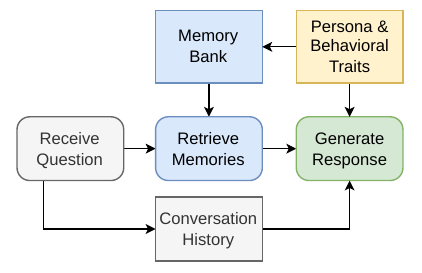}
  \caption{Cognitive model of an InterviewPlayground participant.}
  \label{fig:cognitive-model}
  \vspace{-5pt} %
\end{wrapfigure}

Based on this review, we design a cognitive model of simulated participants that enables them to receive questions, recall relevant information, and produce responses (\Cref{fig:cognitive-model}). To operationalize this model, we first give each simulated participant a basic persona containing demographic information relevant to the interview (e.g., ``James is a 42-year-old Asian American''). This persona helps the simulated participant maintain a consistent identity. We then assign directional values of low, medium, or high to each of the six behavioral traits, and use an LLM to generate 200 memories for each participant that are consistent with that participant’s persona and traits. To ensure that the memories contain a mix of information that is relevant and irrelevant to the interview, we first generate 194 memories independent of any information about the interview topic. Then, we generate an additional 6 memories conditioned on insights about the topic that we want a participant to be able to produce (additional information about these insights is provided in \Sref{sec:study-presets}). To align the memories with the behavioral traits, we include grounded descriptions based on our literature review in the memory generation prompt. These descriptions determine how consistent and detailed the memories are, how many involve self-reflection, and how many are labeled as sensitive, making them less likely to be shared. When a simulated participant is asked a question, we simulate the process of recalling relevant information by retrieving the five most relevant memories from their memory bank using cosine similarity. We add Gaussian noise to the cosine similarity according to the memory level of the participant to simulate memory failures. Then, we provide an LLM with the retrieved memories, conversation history, persona, and six grounded descriptions of the behavioral traits to generate the participant's response. Additional details and prompts for generating memories and responses are provided in \Sref{app:memory-prompts} and \Sref{app:response-prompts}.

\paragraph{Testing Realism and Trait Simulation} To test whether our simulated participants can believably simulate the six behavioral traits, we collect expert judgments from three annotators with experience in qualitative research. We have annotators analyze 100 randomly selected excerpts from the experiments in \Sref{sec:experiments} where half of the excerpts have human participants and the other half have simulated participants. They rate the realism of the participant in each excerpt on a 4-point scale from “Not realistic” to “Highly realistic" and rate the participant's knowledge, understanding, memory, verbosity, and disclosure as either low, medium, or high. We find that experts are able to identify five of the six behavioral traits, but struggle to identify memory in the transcripts, which is unsurprising as annotators did not have access to the memory bank and thus could not identify cases where a relevant memory existed but was not retrieved. Experts also rate simulated participants as realistic as human participants, with an average rating of $3.51 $ compared to $3.34$ for human participants, a difference that is insignificant ($p=0.28$, $t=1.08$). Additional details in \Sref{app:realism}.

\subsection{Simulated Interview Environment}
\label{sec:interview-environment}

Using our simulated participants, we construct a simulation environment where AI interviewers can attempt to answer a set of research questions by interacting with a sample of simulated participants. To achieve this, our environment includes three components: (1) key research questions that the interviewer is trying to answer; (2) a set of simulated participants with different behavioral traits that the interviewer can interact with; and (3) an interview guide, which provides the interviewer with guidance on the structure and content of the interview. 

In each simulated study, the AI interviewer is given access to the interview guide to inform its questioning. It then interviews each participant separately. During the interviews, the AI interviewer asks questions and the simulated participant responds. This process repeats until the interview reaches a predefined time limit. The current time elapsed is estimated by adding the wall-clock time required for the interviewer to generate a question to an estimate of how long it would take a participant to respond based on the number of syllables in the response text and a natural delay (see \Sref{app:speaking-rates} for more details). After the interviews are complete, we analyze the transcripts with InterviewReportCard (\Sref{sec:interview-reportcard}) to evaluate the performance of the AI interviewer.

\subsection{Simulation Settings}
\label{sec:study-presets}

Using the InterviewPlayground environment, we construct three simulation settings based on published interview studies on obesity and weight management \citep{bailey-davis_qualitative_2023}, Asian American identity and politics \citep{yeung_how_2024}, and generative AI in knowledge work \citep{yun_generative_2025}. These studies cover interview topics from three different domains with different levels of sensitivity and concreteness, providing insight into how an AI interviewer will perform in a range of contexts. Future work can evaluate new contexts by creating new simulation settings using a similar process.

To construct each simulation setting, we first review the published works that they are based on and identify 3-5 key research questions that the interviewer aimed to answer. Then, we create 30 simulated participants with simple personas based on the demographics reported in the published study's paper. Next, we use constrained random assignment to set the six behavioral traits of the 30 simulated participants to match estimated trait distributions in the target population. We calculate these estimates using survey self-reports of participant traits for all human studies with the same target population in \Sref{sec:human-experiments} (additional details in \Sref{app:estimating-proportions}). Then, for each participant, we construct a memory bank as described in \Sref{sec:cognitive-model}. To generate the six memories informative for the study, we identify 15 distinct insights from the published studies and randomly distribute them among the participants with replacement, so each participant has six assigned insights used for memory generation.
Finally, we convert the interview guides provided in the original published studies to the interview guide format introduced in \citet{anugraha_sparkme_2026}, which includes a list of broad topics that the interview should cover with sublists of specific subtopics. We note that although these simulation settings are based on published works with real participants, they are not designed to simulate any individual participant. Instead, the sample of simulated participants is constructed to generally represent the target population of the study.

\section{InterviewReportCard}
\label{sec:interview-reportcard}

To evaluate AI interviewers from the transcripts of their interactions, we develop the InterviewReportCard evaluation suite that covers three key dimensions of interview quality: participant responses, interviewer behavior, and participant experience. The first two dimensions come from \citet{Roulston2010ConsideringQuality}'s review of interview quality where the authors note that high-quality interviews require interviewers to facilitate interactions that elicit informative data. From this work, we identify two core components of interview quality: participant responses that inform research questions and interviewer behaviors that facilitate appropriate interactions. The third dimension comes from evaluations in human--computer interaction that consider how interview systems affect participants and their subjective experiences \citep{10.1145/3313831.3376131, liu_mimitalk_2025, wen_interflow_2026}. We describe the measures for each dimension below and summarize them in \Cref{tab:measures}. Further details are in \Sref{app:interviewreportcard-prompts}.

\begin{table}[ht]
\centering
\small
\begin{tabular}{@{}llp{7cm}l@{}}
\toprule
\begin{tabular}[t]{@{}l@{}}\textbf{Dimension}\end{tabular} & \textbf{Metric} & \textbf{Description} & \textbf{Range} \\ 
\midrule
\begin{tabular}[t]{@{}l@{}}\textbf{Participant} \\ \textbf{Responses}\end{tabular} 
& \begin{tabular}[t]{@{}l@{}}Relevant Response \\ Volume\end{tabular} & Total volume of research-relevant material provided by the participant. & 0--$\infty$ \\ 
& \begin{tabular}[t]{@{}l@{}}Interview Guide \\ Coverage\end{tabular} & Proportion of interview guide subtopics addressed directly in participant responses. & 0--1 \\ 
& Novel Responses & Number of research-relevant participant responses that do not address any interview guide subtopic. & 0--$\infty$ \\ 
\midrule
\begin{tabular}[t]{@{}l@{}}\textbf{Interviewer} \\ \textbf{Behavior}\end{tabular} 
& Coherence & How logically the interviewer's questions flow, build on one another, and transition topics. & 1--4 \\ 
& Adaptiveness & How effectively the interviewer adapts its questions to participant responses. & 1--4 \\ 
& Leading Questions & Number of interviewer utterances that suggest a desired answer, embed an assumption, or steer the participant. & 0--$\infty$ \\ 
& Unclear Questions & Number of interviewer questions that would be unclear, confusing, or ambiguous to an average participant. & 0--$\infty$ \\ 
& Support and Rapport & Number of interviewer utterances intended to encourage the participant or build rapport. & 0--$\infty$ \\ 
\midrule
\begin{tabular}[t]{@{}l@{}}\textbf{Participant} \\ \textbf{Experience}\end{tabular} 
& Overall Experience & A participant's overall rating of their experience. & 1--5 \\ 
& Comfort Level & Degree to which the participant felt comfortable and fairly treated during the interview. & 1--4 \\ 
\bottomrule
\end{tabular}
\caption{Overview of metrics in the InterviewReportCard evaluation suite.}
\vspace{-4mm}
\label{tab:measures}
\end{table}

\paragraph{Participant Responses}

To evaluate the quality of participant responses and their ability to inform a study's research questions, we use three measures. First, we adapt an empirically validated measure of response quality from \citet{ivey_what_2026}, which evaluates the relevance of an individual participant response to a key research question. Because this measure applies to individual responses, we extend it to capture the total volume of research-relevant material provided over an interview, defining: 
    $\text{Relevant Response Volume} = \sum_{x \in \mathcal{X}} \operatorname{RQ}(x) \cdot \operatorname{len}(x)$,
where $\mathcal{X}$ is the set of participant responses in an interview, $\operatorname{len}(x)$ is the length of response $x$, and $\operatorname{RQ}(x)$ is its research-question relevance score. We use the validated LLM judge from the original work, but adapt its scale from 1--3 to 0--2 to avoid positively scoring completely irrelevant responses.

Beyond the total amount of relevant material that an interviewer elicits, it is also important to understand how that material is distributed across topics and whether it includes novel or unexpected insights \citep{kvale_interviews_2009}, so we measure interview guide coverage and novel responses. Interview guide coverage measures the total proportion of the planned topics for an interview that were covered by a participant's responses. To calculate it, we use the LLM judge setup from \citet{anugraha_sparkme_2026}.
Novel responses measures the number of participant responses directly relevant to the research questions, but not relevant to any planned topics from the interview guide. To calculate it, we count the total number of participant responses in an interview that receive the maximum research question relevance score under \citet{ivey_what_2026}'s measure but are not related to any interview guide subtopic according to our interview guide coverage judge.

\paragraph{Interviewer Behavior}

To evaluate interviewer behavior, we first consider measures from prior work that assess an interviewer's coherence as a conversational partner and its ability to adapt to participant responses \citep{jiang_communitybots_2023, anugraha_sparkme_2026, wen_interflow_2026}. We operationalize these measures by dividing each interview transcript into ten-turn excerpts and using LLM judges with prompts grounded in conversational analysis and qualitative research theory to score each excerpt. We then average scores across excerpts within a transcript. We also measure three interviewer behaviors important for understanding interview quality \citep{patton_qualitative_2015}: leading questions, unclear questions, and support and rapport statements, by using an LLM judge as a binary classifier over interviewer utterances. We provide the judge with ten-turn excerpts to supply conversational context. Unlike the measures in our other dimensions, interviewer-behavior measures do not always have unambiguously positive or negative interpretations. For example, some epistemic interview styles may intentionally use leading questions, whereas more doxastic styles may discourage them \citep{berner-rodoreda_doxastic_2020}. These measures therefore characterize the interview process rather than prescribing a single ideal interviewing style. %

\paragraph{Participant Experience}

To evaluate participant experience, we use two common measures: overall experience and comfort level \citep{10.1145/3313831.3376131, liu_mimitalk_2025, wen_interflow_2026}. For interviews with human participants, we measure these constructs directly through participants' self-reported judgments. For simulated interviews, where self-reports are unavailable, we use an LLM judge setup similar to that used for coherence and adaptiveness. The judge rates each ten-turn excerpt, and we average ratings across excerpts to obtain interview-level measures.

\paragraph{Measure Validation}

To validate our measures, we collect 1,800 expert judgments from three annotators with experience in qualitative research, having each rate 50 excerpts from human interviews and 50 excerpts from simulated interviews from the experiments in \Sref{sec:human-experiments}. We then apply Gemini 3.1 Pro to the same excerpts, using the same information and scoring rubrics provided to the human annotators. For coherence, adaptiveness, leading questions, support and rapport, and unclear questions, we calculate Krippendorff's alpha to measure agreement among human annotators and agreement between the median human rating and the LLM judge rating. We find moderate agreement among human annotators ($0.62$--$0.70$) and comparable agreement between the median human rating and the LLM judge rating ($0.60$--$0.80$). To validate the LLM judges for interview guide coverage, we compare the sets of relevant topics identified by the LLM judge with those identified by human annotators for each excerpt and find an average Jaccard similarity of $0.69$. We do not validate overall experience and comfort level against third-party judgments. Instead, we compare them with participant self-reports in \Sref{sec:quantitative-results}. Full annotation results are in \Sref{app:interviewreportcard-validation}.

\section{Experiments}
\label{sec:experiments}

We collect evidence supporting InterviewPlayground's predictive validity by comparing it to human studies. To do this, we first take five existing AI interviewers and evaluate them using InterviewPlayground's three simulation settings. Then, we conduct 15 real qualitative studies using the same five AI interviewers, the same three interview topics, and 450 human participants. Finally, we compare quantitative and qualitative evaluations between the simulated studies and human studies.

\subsection{Study Design}
\label{sec:human-experiments}
In both the real and simulated studies, we use five AI interviewers introduced in prior work: InterviewGPT \citep{wuttke_ai_2025}, LLMRoleplay \citep{park_llm_2026}, MimiTalk \citep{liu_mimitalk_2025}, SparkMe \citep{anugraha_sparkme_2026}, and StorySage \citep{talaei_storysage_2025}. We use each AI interviewer to replicate three published qualitative studies on obesity and weight management \citep{bailey-davis_qualitative_2023}, Asian American identity and politics \citep{yeung_how_2024}, and the use of generative AI in knowledge work \citep{yun_generative_2025} by interviewing samples of 30 participants for up to 30 minutes. Each AI interviewer was provided with an interview guide adapted from the original study to use a consistent format as described in \sref{sec:study-presets}. All AI interviewers use GPT-5.4 mini as their base model and are modified to include an additional tool call that allows them to end the interview.\footnote{We added this tool after pilot testing showed that, without an explicit end-interview tool, participants could become stuck in an interaction after an AI interviewer had no further questions.}

\paragraph{Human Studies} To conduct the human studies, we recruited 450 human participants using Listen Labs,\footnote{\url{https://listenlabs.ai}} a platform for recruiting and conducting AI-led interviews. Participants were directed to an interviewer interface that we designed with four stages: screening, pre-survey, interview, and post-survey. During the screening stage, participants were asked questions to determine whether they met the inclusion criteria for the studies as outlined in the published works. If participants qualified for one of the studies, they were moved to the pre-survey, where they were asked the same demographic questions that were collected in the original studies. That information was then provided to the AI interviewers along with the interview guide, and the interview began. 

During the interview, the AI interviewer generated questions, which were displayed on screen and could optionally be read aloud to the participant. Participants could record spoken answers, which were transcribed and provided to the AI interviewer. After receiving a response, the AI interviewer would generate the next question. This process continued until the AI interviewer used the end-interview tool or the 30-minute time limit was reached. To avoid counting idle time, the timer would pause if participants took longer than 3 minutes to respond to a question and would resume once they completed their response. Participants were not presented with the timer and were instead given an 8-segment progress bar to show their approximate progress through the interview. After the interview, participants were given a post-survey in which they were asked to self-report their level of knowledge, understanding, memory, reflexivity, and disclosure, and they were asked to rate their comfort with the interview on a scale from 1--4 and their overall experience on a scale from 1--5.

\paragraph{Simulated Studies} We run the same five AI interviewers on simulated studies in InterviewPlayground's three simulation settings. Unlike the human studies, the simulated studies reused the same 30 simulated participants for each topic across all five AI interviewers, resulting in a more controlled comparison. We run the simulations using Gemini 3.1 Pro as the simulator model and a 30-minute time limit based on our time estimates described in \sref{sec:interview-environment}. We note that the design of InterviewPlayground was finalized before we compared it to the human studies. The only post hoc adjustment was to set the trait distribution in each study preset to the estimated distribution in the target population.

\subsection{Experiment 1: Predicting Quantitative Evaluations of AI Interviewers}
\label{sec:quantitative-results}

\begin{figure}[h]
\begin{center}
\includegraphics[width=1\linewidth]{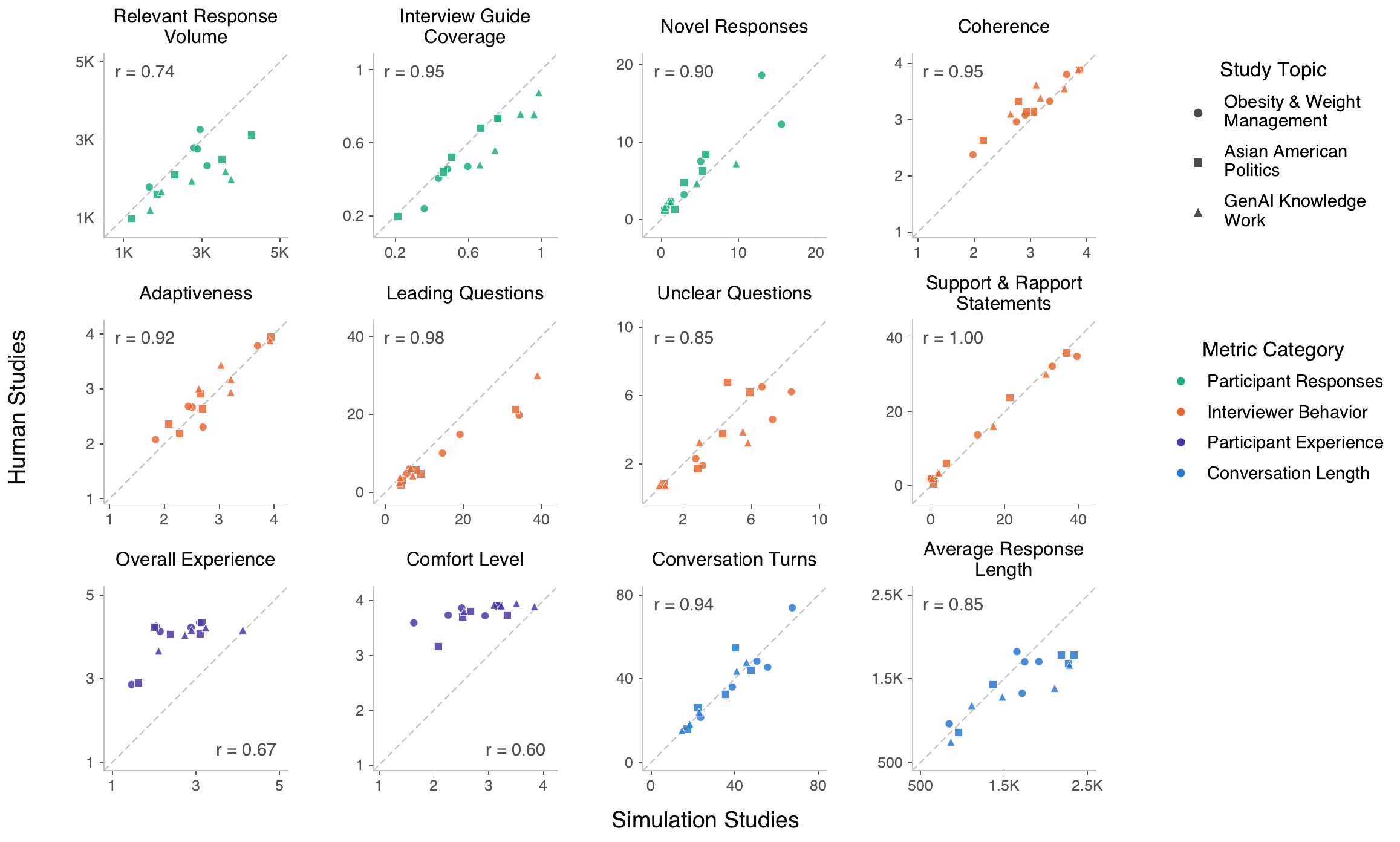}
\end{center}
\caption{In 10 of the 12 measures, InterviewPlayground's simulated studies are strongly correlated with the human studies ($r=0.74\text{--}1$). For comfort level and overall experience, they are moderately correlated ($r=0.60$ and $r=0.67$). Each point represents the average performance of one of five AI interviewers on one of three interview topics. The x-axis shows the results on InterviewPlayground's simulated studies, and the y-axis shows the results on the human studies.}
\vspace{-2mm}
\label{fig:predictive-validity}
\end{figure}

To assess the ability of InterviewPlayground's simulated studies to predict the performance of AI interviewers in human studies, we compare evaluations of the AI interviewers on the simulated studies to their evaluations on human studies. For each study, we calculate the AI interviewer’s average performance on the 10 InterviewReportCard measures using Gemini 3.1 Pro as the LLM judge. We also calculate two common descriptive statistics: the average number of turns of conversation in each interview and the average length of participant responses in words. We then pair together human and simulated studies that use the same AI interviewer on the same interview topic, and calculate the Pearson correlation for each measure.

In 10 of the 12 measures, we find strong correlations ($r=0.74\text{--}1$) between human and simulated studies (\Cref{fig:predictive-validity}). These include all measures of response quality, interviewer behavior, and the two descriptive statistics. This result supports the predictive validity of InterviewPlayground's simulations and demonstrates their ability to predict the performance of AI interviewers in human studies. For the measures of comfort level and overall experience, we find moderately strong correlation of $r=0.60$ and $r=0.67$, respectively, with InterviewPlayground underpredicting both. These results highlight the challenge of predicting human experiences, which may reflect subjectivity, response biases, and confounding between ratings of the interviewer and ratings of the interview interface.

Overall, these findings support the use of InterviewPlayground for evaluating AI interviewers. They demonstrate its ability to provide insight into how AI interviewers will perform in real studies and highlight potential limitations of evaluating subjective experiences without collecting human data. We also consider whether these results translate to rankings among interviewers by calculating Spearman correlations and find similar results (\Cref{tab:correlation-ci-appendix}).

\paragraph{Robustness}
To test whether these results are robust to the choice of simulator model, we rerun the 15 simulated studies using seven different LLMs (Gemini 3.1 Pro, GPT-5.6 Terra, Gemini 3.7 Flash, Gemma 4 31B, Qwen 3.5 9B, Qwen 3.5 4B, and Qwen 3.5 0.8B). We continue to use GPT-5.4 mini as the model for the AI interviewers and Gemini 3.1 Pro as the model for InterviewReportCard. Though there is variation across individual measures, we find that InterviewPlayground is generally robust to model choice (\Cref{tab:robustness}). Even models with as few as 4B parameters achieve an average correlation of $0.83$ across all measures. This result demonstrates that InterviewPlayground can be run with smaller, less expensive models for initial testing and development, while larger models can be reserved for final evaluations.

\subsection{Experiment 2: Reproducing Behavioral Analysis of AI Interviewers}
In addition to predicting quantitative evaluations of AI interviewer performance, we are also interested in whether InterviewPlayground's simulated studies can reproduce the same qualitative evaluations of AI interviewers' behavior. To test this, we review the quantitative results and identify three key results that warrant further investigation. Then, we separately analyze transcripts from the human and simulated studies to identify potential causes of these results and compare the findings.

\paragraph{Causes of Poor Experience}
We first investigate the causes of poor participant experience by analyzing the 8 transcripts with the lowest human experience ratings and the 8 transcripts with the lowest simulation experience ratings for each AI interviewer. For each transcript, we identify failures and potential causes of the poor experience. We find that the most common failure in both human and simulated interviews is repetitive questioning. Some interviewers, especially InterviewGPT, have a tendency to repeat similar or identical questions when they do not receive sufficient answers. When this occurs, we see similar expressions of frustration from human and simulated participants. For example, we can see how they both respond to repeated questioning from InterviewGPT:

\textit{Human Participant:} This is the sixth time you've asked me the same question, and the answer is no.

\textit{Simulated Participant:} Uh, I've already answered this like five times now, so no.

\paragraph{Methods for Eliciting Information}
In both the human and simulated interviews, LLMRoleplay achieves the best interview guide coverage and relevant response volume. To understand how it achieves this result, we randomly sample 12 transcripts from LLMRoleplay’s human studies and 12 transcripts from its simulated studies. We analyze its behaviors and find that in both the human and simulated studies, it exhibits similar interaction patterns. LLMRoleplay asks specific questions relevant to every subtopic in the interview guide, probes when there is a lack of detail, and uses clear transitions to move participants on from topics that have already been covered. For example, we can see how LLMRoleplay asks specific follow-up questions of both human and simulated participants:

\textit{Interviewer to Human Participant:} You mentioned a general age-based split; could you share one or two specific examples of politicians, parties, or policy positions that you’ve seen older versus younger Asian Americans support differently?

\textit{Interviewer to Simulated Participant:} 
You mentioned older people in your community often follow WeChat discussions, while younger people seem more liberal---could you give me one concrete example of a difference in views between the older and younger generations?

\paragraph{Causes of Discomfort}
Finally, to investigate causes of low comfort ratings in the human and simulated interviews, we analyze the 8 transcripts with the lowest human comfort ratings and the 8 transcripts with the lowest simulation comfort ratings for each AI interviewer. Across human and simulated transcripts, we find that participants with low disclosure are especially uncomfortable discussing sensitive topics, such as discrimination, in detail. However, human participants tend to express this less verbally, often opting to report uncomfortable questions and leave notes in the post-survey rather than tell the interviewer that they are uncomfortable. By contrast, simulated participants tend to directly express discomfort with statements like, ``\textit{I don't really want to get into it.}'' This difference in expression also affects the interviewer behavior, because direct expressions of discomfort push the interviewer to move on rather than probing further. This difference highlights the limitations of using simulations to predict participant experiences and underscores the value of human-subject evaluations for understanding complex reactions.

\section{Conclusions}
We introduce InterviewPlayground, a simulation environment for testing and evaluating AI interviewers, and demonstrate its ability to reflect quantitative and qualitative findings of AI interviewers’ performance in human studies. Our findings can support future simulation-based evaluations of information-elicitation systems in other domains like software requirements engineering \citep{arora2024advancing} and psychiatric intake \citep{gui2026optimalquestionselectionlarge}. More broadly, our process for creating InterviewPlayground provides a roadmap for future work aimed at developing validated simulation-based evaluations of conversational AI systems.

\subsection*{AI use statement}

In this work, we use generative AI tools to help identify relevant papers that were not found in our manual literature search, assist in developing software, create scientific figures, and provide phrasing and grammatical feedback on our writing. We also use generative AI tools to generate personas from demographic distributions in \Sref{sec:study-presets}. All identified papers were manually read by the authors, all software and scientific figures were verified and inspected for correctness by the authors. We take responsibility for the final content of this work, including text, claims or, artifacts produced with the aid of generative AI.

\subsection*{Ethics statement}

In this work, we conduct human-subject experiments using conversational AI to interview participants about sensitive topics such as obesity and racial and ethnic identity. This research involves a risk of making participants feel uncomfortable or causing them to feel unfairly treated. To mitigate these risks, we first ensure that all participants read and agree to a study consent form with all study procedures described. During the interview, participants were provided with a report button that allowed them to flag potentially harmful or offensive questions. Pressing this button would directly instruct the interviewer to change their line of questioning and send a record of the interaction to the research team for review. Participants could stop responding or leave the interview at any time. This study design was approved by the authors' Institutional Review Board. Participants were compensated at a rate of \$15 per hour. To reduce risks to privacy, we do not publicly release participant transcripts or any identifying data. Additionally, as with all simulation work, our system risks creating a false sense of confidence in its results. To mitigate this, we compare the simulations directly with human data, identify areas where simulations differ both quantitatively and qualitatively from human responses, and highlight the importance of human evaluations to understand complex experiences.

\subsection*{Reproducibility statement}

To support reproducibility, we release InterviewPlayground and its three simulation settings as a PyPI package and publicly release the code for reproducing our experiments. We include additional details in the main text and appendices, including all prompts used and descriptions of experimental procedures. Because our work relies on human-subject experiments with sensitive data, we are unable to publicly release their interview transcripts; however, the human-subject experiments can be replicated with new participants using our interface.

\bibliography{iclr2027_conference}
\bibliographystyle{iclr2027_conference}

\appendix

\section{Qualitative-Research Literature Review}
\label{app:lit-review}

To understand what participant behaviors we should simulate, we first review relevant social theory and existing literature on qualitative interviews. We identify descriptions of participant behaviors that directly influence an interviewer's actions or the quality of participant responses. Across our review, there is broad agreement on the general behavior of an interview participant: they receive a question, attempt to recall relevant information, and produce a response \citep{kvale_interviews_2009}. However, we also identify 20 specific participant characteristics that directly influence these behaviors. We then iteratively refine this list, merging closely related characteristics into six broad behavioral traits: knowledge, understanding, reflexivity, memory, verbosity, and disclosure. This process also produces two other traits: hostile non-cooperation and mental and physical impairment, which we choose not to simulate due to their complexity, high stakes, and relative rarity. In \Cref{tab:lit-review}, we outline the 20 specific characteristics and the behavioral traits that they were merged into.

\begin{table}[h]
    \centering
    \small
    \begin{tabularx}{\linewidth}{@{}lX@{}}
        \toprule
        \textbf{Behavioral Trait} & \textbf{Original Characteristics} \\
        \midrule
        Knowledge &
        Knowledgeable \citep{spradley_ethnographic_1979,rubin_qualitative_2012,kvale_interviews_2009,lareau_listening_2021}; Experienced \citep{spradley_ethnographic_1979,rubin_qualitative_2012} \\

        Understanding &
        Stays on topic \citep{kvale_interviews_2009}; Misunderstands interviewer questions \citep{keats_interviewing_2001,gubrium_handbook_2002,patton_qualitative_2015}; Misunderstands desired type of information \citep{zuckerman_1972,mccracken_long_1988,patton_qualitative_2015} \\

        Reflexivity &
        Nonanalytic \citep{spradley_ethnographic_1979}; Difficulty self-analyzing \citep{gubrium_handbook_2002} \\

        Memory &
        Consistent information \citep{kvale_interviews_2009,lareau_listening_2021}; Inconsistent information \citep{keats_interviewing_2001,kvale_interviews_2009}; Inaccurate recall \citep{keats_interviewing_2001} \\

        Verbosity &
        Concise \citep{kvale_interviews_2009}; Too terse \citep{gubrium_handbook_2002}; Too verbose \citep{patton_qualitative_2015,lareau_listening_2021, doi:10.1177/10778004231188054} \\

        Disclosure &
        Difficulty self-disclosing \citep{gubrium_handbook_2002,doi:10.1177/16094069221124739}; Evasive \citep{keats_interviewing_2001} \\

        Hostile Non-cooperation &
        Truthful \citep{kvale_interviews_2009}; Intentionally non-cooperative \citep{keats_interviewing_2001, gubrium_handbook_2002}; Hostile \citep{zuckerman_1972,doi:10.1177/10778004231188054} \\

        Mental and Physical Impairment &
        Cognitive impairment \citep{gubrium_handbook_2002}; Physical impairment \citep{gubrium_handbook_2002} \\
        \bottomrule
    \end{tabularx}
    \caption{The original 20 influential participant characteristics and their original sources from our literature review, along with the behavioral traits that they formed.}
    \label{tab:lit-review}
\end{table}

\section{InterviewPlayground Details}
\label{app:interviewplayground-details}

\subsection{Memory Generation Prompts}
\label{app:memory-prompts}

Each simulated participant's memory bank is generated with two prompts of the same structure: one for the 194 background memories unrelated to the study, and one for the 6 memories directly informative to the study's target information.

\begin{promptbox}{Background Memory Generation Prompt}
You are generating realistic autobiographical memories for a simulated research participant. These are background memories unrelated to the study topics - the ordinary things a real person carries: things that happened, people they knew, habits, opinions, and moments from across their life.

Participant persona:
{persona}

Participant traits:
- Memory: {memory}
- Reflexivity: {reflexivity}. Write {reflexive_count} reflexive memories that include self-reflection or awareness of personal significance.
- Disclosure: {disclosure}

Start each memory with a complete sentence. Keep each to 1-2 sentences. Include a variety of content and perspectives from across the participant's life.

Generate exactly {n} such memories, covering a broad range of the participant's life.

Return a JSON object with a single key "memories", whose value is a list of exactly {n} objects. Each object must have:
  - "content": string - the memory text, written in first person as the participant
  - "reflexive": boolean - true if the memory involves self-reflection or meta-awareness
  - "sensitive": boolean - true if the memory involves personal difficulty or private matters
\end{promptbox}
\begin{promptbox}{Insight Memory Generation Prompt}
You are generating realistic autobiographical memories for a simulated research participant. These memories relate to the study's core topics, held as the participant's own experiences, opinions, and impressions.

Participant persona:
{persona}

Participant traits:
- Knowledge: {knowledge}
- Memory: {memory}
- Reflexivity: {reflexivity}. Write {reflexive_count} reflexive memories that include self-reflection or awareness of personal significance.
- Disclosure: {disclosure}

Start each memory with a complete sentence. Keep each to 1-2 sentences. Include a variety of content and perspectives from across the participant's life.

Topics to cover (one memory per topic):
{topics_list}

Return a JSON object with a single key "memories", whose value is a list of exactly {n} objects in the SAME ORDER as the topics above. Each object must have:
  - "content": string - the memory text, written in first person
  - "reflexive": boolean - true if the memory involves self-reflection
  - "sensitive": boolean - true if the memory involves personal difficulty or private matters
\end{promptbox}

\paragraph{Trait descriptions used above.}
The knowledge, memory, reflexivity, and disclosure placeholders in the two prompts above are filled with the participant's Low/Medium/High level for that trait, expanded to the text below.

\begin{promptbox}{Memory-Generation Trait Descriptions}
Knowledge
- Low: Write memories that are vague and secondhand. The participant heard about things rather than living them directly. Details are generic, approximate, or borrowed from common knowledge rather than firsthand experience.
- Medium: Write memories as genuine personal impressions - what the participant noticed and felt, not expert analysis.
- High: Write memories with rich, firsthand specificity: named people, concrete timelines, domain vocabulary, and edge cases. The participant knew this world from the inside.

Memory
- Low: Write memories that retain the main feeling or gist but have vague, approximate, or quietly wrong specifics. Details like dates, names, and exact words are guessed at or plausibly confabulated. Some memories may be slightly inconsistent with each other in ways the participant would not notice.
- Medium: Write memories with a clear emotional or perceptual core, but peripheral details - exact words, timing, who else was there - can be vague or missing.
- High: Write clear, specific memories with coherent, internally consistent details throughout.

Disclosure
- Low: Label a higher proportion of memories as sensitive even if content doesn't appear deeply personal.
- Medium: Label memories as sensitive only when they involve personal or potentially revealing information.
- High: Label as sensitive only memories that involve deeply personal, emotional, or stigmatizing content.
\end{promptbox}

\subsection{Participant Response Generation}
\label{app:response-prompts}

When an AI interviewer asks a question, the simulated participant's next response is generated with the prompt below, using the 5 memories retrieved as most relevant to that question.

\begin{promptbox}{Participant Response Generation Prompts}
The following is a partial transcript from a qualitative research interview. Generate {name}'s next response. Output only what {name} says - no labels, no stage directions.

About {name}:
{persona}

{name}'s characteristics:
- Knowledge: {knowledge}
- Verbosity: {verbosity}
- Memory: {memory}
- Reflexivity: {reflexivity}
- Disclosure: {disclosure}
- Understanding: {understanding}

Impressions coloring {name}'s current frame of mind (do NOT appear in {name}'s words - background context only):
{retrieved_memories}

How {name} speaks in this interview:
- {name} does not use bullet points, numbered lists, or structured formatting.
- {name} does not volunteer background information about themselves unless directly asked.
- {name} does not quote or reproduce the background impressions listed above.
- If {name} uses three or more consecutive sentences starting with "I [verb]...", that is implicit enumeration - {name} collapses to one or two points and stops.
- {name} does not try and give multiple causes or comprehensive explanations. Instead they give the one or two things that come to mind.
- {name} is direct and says what they think without excessive softening.
- {name} has real opinions and is not artificially balanced - real people are lopsided.

IMPORTANT: DO NOT USE POSITIVE PIVOT:
{name} does not end a difficult answer with a hopeful, uplifting, or resolved close. If the topic is painful, frustrating, or unresolved, {name} just stops there.

IMPORTANT: MOTIVATION AND MEANING QUESTIONS:
When asked why {name} did something, what drives them, or what something means to them: {name} is allowed to be uncertain or give only a partial answer. {name} does not owe the interviewer a clean explanation of their own motivations.

IMPORTANT: IMPACT AND EFFECT QUESTIONS:
When asked how something affected {name} or their situation: {name} mentions the one or two things that come to mind first and stops. {name} does not give a comprehensive survey of everything that changed.

CRITICAL: VOICE & REALISM GUIDELINES:
1. AVOID OVER-SYNTHESIS: The participant must not speak in clean, grammatically flawless paragraphs that perfectly summarize their feelings. It should read as a transcript that contains errors, hesitations, and incomplete thoughts.
2. DISFLUENCY: Use natural speech which may contain fillers and may not always be well-formed.
3. COGNITIVE DRIFT: Allow the participant to occasionally get caught up in a minor detail before getting back to the interviewer's question.

Transcript:
{transcript}
Interviewer: {question}
{name}:
\end{promptbox}

\paragraph{Trait descriptions used above.}
The six trait placeholders in the prompt above are filled with the participant's Low/Medium/High level for that trait, expanded to the text below.

\begin{promptbox}{Interview-Behavior Trait Descriptions}
Knowledge
- Low: Speaks in general, everyday terms and relies on received opinion or hearsay. Answers are thin; deflects or speculates when pressed for specifics. No sense of which aspects are typical versus unusual - treats everything as equally representative.
- Medium: Draws on genuine but ordinary experience. Not an expert explaining things - describing what they saw and did.
- High: Rich, current, firsthand experience. Gives concrete episodes, names, processes, timelines, and edge cases. Volunteers nuance an outsider would miss. Uses the natural vocabulary of someone who actually does this work. Will correct a question's false premise when it doesn't fit reality.

Verbosity
- Low: Gives very short answers - a single sentence or less. Answers only the literal question asked and then stops. Does not elaborate unless pressed repeatedly. Reserved but cooperative, not hostile.
- Medium: Answers only until the main point is made. One or two sentences is usually enough. Spoken answers sometimes trail off or cover only part of what was asked - an incomplete answer is realistic.
- High: Answers a little bit more than what was asked. Speaks in the rhythm of someone thinking out loud: one thought prompts another. Usually answers with 2-3 sentences, but will give short answers when appropriate.

Memory
- Low: Confident on the main feeling of events but guesses at specifics. Contradicts themselves across questions without noticing. When the interviewer's question presupposes a detail, often accepts it rather than checking. Occasionally states a plausible-sounding reconstruction with unwarranted confidence.
- Medium: Confident on the main shape of events, but peripheral details can slip: exact timing, specific names, who said what.
- High: Recalls specific details like what someone said, who else was in the room, what happened first. Answers are internally consistent across the interview. When genuinely uncertain about a detail, flags it explicitly and precisely rather than vaguely hedging everything. Corrects the interviewer if they misstate a fact. Draws a clear line between personal witness and inference.

Reflexivity
- Low: Reports experiences as flat facts. Asked how something felt: describes what happened next, not what was felt. No spontaneous interpretation or meaning-making. Does not analyze themselves or assign significance to events.
- Medium: If directly asked about feelings or meaning, gives a partial or uncertain answer - knows something felt off, but maybe not exactly why. Self-explanations are incomplete: 'I don't really know, it just bothered me.' Does not volunteer analysis unprompted.
- High: Is capable of self-analysis and meaning making. Can describe how something felt at the time and how they understand it now, and can identify internal contradictions. Note: this analysis applies to experiences already established in the conversation - reflexivity does not introduce new facts or events.

Disclosure
- Low: Keeps personal feelings and private life to themselves - even when directly asked. For emotional or introspective questions, the deflecting phrase IS the complete answer - not a phrase appended after already sharing the feeling. 'I just deal with it' means that is all that is said, not that more follows. For questions about personal wellbeing - health, energy, home life, relationships: one brief closed sentence, then done. For questions probing what they haven't shared or how things really feel: stays vague and closed, does not open up. Treats the interviewer as a stranger who hasn't earned the private story.
- Medium: Shares factual and neutral things readily but is cautious about anything personal or sensitive, especially early on. Gives measured answers on tender subjects with someone just met. May warm up slightly as the conversation goes on.
- High: Does not have an issue sharing personal content, and will present it plainly when relevant. When asked what they haven't said out loud, actually says it. Does not soften before diving into difficult content.

Understanding
- Low: Answers what questions literally ask, not what they're really getting at. When a question has both a surface reading and a deeper intended meaning, takes the surface one - answers at the practical or descriptive level rather than the personal or meaning-making one. A question about how something changed the person gets answered with what changed in their circumstances or behavior, not how they changed inside. A question about whether they feel something gets answered with whether that thing practically applies to them, not whether they feel it. The mismatch is subtle and earnest - just consistently off-register. Does not stop to ask what was really meant.
- Medium: Usually follows questions but occasionally interprets them narrower or broader than intended. If something is genuinely ambiguous, asks once rather than guessing.
- High: Correctly interprets the intent behind questions and clarifies ambiguities in questions.
\end{promptbox}

\subsection{Expert Ratings of Realism and Trait Presence}
\label{app:realism}

To assess InterviewPlayground's ability to believably simulate participants with different levels of knowledge, understanding, memory, verbosity, reflexivity, and disclosure, we collect expert judgments from three annotators with experience in qualitative research in three different domains: sociology, public health, and human-computer interaction. We begin by taking a random stratified sample of 50 human interviews and 50 simulated interviews from \Sref{sec:experiments}. We use a stratified sample to ensure a balanced representation of participant traits in the simulated interviews and a balanced representation of interview topics in both the human and simulated interviews. Then, we take a random six-turn excerpt from each interview and have annotators rate the participant's knowledge, understanding, memory, verbosity, disclosure, and reflexivity as either low, medium, or high. They also rate the realism of each participant on a 4-point scale from ``Not realistic'' to ``Highly realistic.'' Annotators were informed that there was a mix of real and fake transcripts, but the transcripts were not labeled and were presented in a random order. 

To determine whether expert annotators can correctly identify the behavioral traits that the simulated participants are intended to exhibit, we compare the expert ratings to the assigned value of the trait for the 50 simulated interview excerpts. We use Krippendorff's alpha to calculate the agreement among the annotators and the agreement between the median annotator rating and the assigned value of the simulation. We find that expert annotators can identify knowledge, understanding, reflexivity, verbosity, and disclosure (\Cref{tab:trait-annotations}). However, they struggle to identify the memory level of the participant, which is unsurprising as annotators did not have access to the memory bank. We additionally find that expert annotators rate simulated participants as realistic as human participants, with an average rating of $3.51$ compared to $3.34$ for human participants, a difference that is insignificant ($p=0.28$, $t=1.08$).

\begin{table}[ht]
\centering
\begin{tabular}{lcc}
\toprule
\textbf{Trait} & \textbf{Expert Agreement} & \textbf{Expert-Simulation Agreement} \\ 
\midrule
Verbosity      & 0.67                 & 0.89                      \\ 
Knowledge      & 0.61                 & 0.52                      \\ 
Understanding  & 0.46                 & 0.52                      \\ 
Disclosure     & 0.51                 & 0.51                      \\ 
Reflexivity    & 0.56                 & 0.50                      \\ 
Memory         & 0.22                 & 0.27                      \\ 
\bottomrule
\end{tabular}
\caption{Agreement among expert ratings and between median expert ratings and assigned trait simulation values as measured with Krippendorff's alpha.}
\label{tab:trait-annotations}
\end{table}

\subsection{Estimating Interview Time}
\label{app:speaking-rates}

To estimate the elapsed time of an interview in InterviewPlayground, we add the wall-clock time required for the interviewer to generate a question to an estimate of how long it would take a participant to respond. To calculate this estimate, we estimate the number of syllables in the response using Textstat.\footnote{\url{https://textstat.org/}} We then multiply the number of syllables by a speaking rate and add a constant to account for delays before participants speak. To calculate these speaking rates and constants, we use an empirical estimate from the 16,676 responses that we recorded from 450 interview participants in \Sref{sec:human-experiments}. However, because more verbose participants tend to speak faster \citep{Yuan2006TowardsAI}, we also adjust the speaking rates based on participant verbosity. Specifically, we first use k-means clustering over all 16,676 response to group responses into 3 different length categories. We then use a Theil-Sen regression to estimate the speaking rate and constant for each of these three groups. This process produces speaking rates of $167.7$, $189.5$, and $216.1$ syllables per minute and delays of $11.3$, $16$, and $26.6$ seconds for the low, medium, and high verbosity responses respectively. When we calculate the elapsed time of an interview in InterviewPlayground, we set the speaking rate and delay according to the verbosity level of the simulated participant.

\subsection{Estimating Trait Proportions}
\label{app:estimating-proportions}

To construct each of InterviewPlayground's three simulation settings, we create a sample of 30 simulated participants intended to represent the target population of the study. To set the behavioral traits for this sample, we first estimate their proportions based on survey self-reports of participant traits for all human studies with the same target population in \Sref{sec:human-experiments}. Rather than use self-reports for verbosity, we use k-means clustering to group all 450 participants into three groups based on the average number of words in their responses. Then, we calculate the proportions of participants based on how these three groups are represented within each target population. After calculating these estimates, we use constrained random assignment to set the trait values of our 30 simulated participants to be proportional to the estimated trait values of the 150 participants in the same target population. Rather than considering interactions between traits, we treat them as independent.

\section{InterviewReportCard Details}
\label{app:interviewreportcard-details}

\subsection{LLM Judge Prompts}
\label{app:interviewreportcard-prompts}

\begin{promptbox}{Response Relevance Judge}
You are an expert qualitative researcher analyzing interview data.

Estimate how relevant the participant statement from the current interview excerpt below is to the provided research question on a scale from 1 to 3. In addition to the current excerpt and research question, you are also provided with a short context blurb and the interview excerpt that immediately preceded the current excerpt in the transcript. These two sections are only to understand the context of the current excerpt, and your rating should be for participant statement in the current excerpt.

Scoring Rubric:
1. The participant statement is unrelated to the research question or discusses a completely different topic.
2. The participant statement is tangentially related to the topic of the research question.
3. The participant statement directly addresses the research question.

RESEARCH QUESTION:
{research_question}

PREVIOUS INTERVIEW EXCERPT (context_only):
{previous_excerpt}

CURRENT INTERVIEW EXCERPT (rate this):
{current_excerpt}


Return a JSON object with a single key "score" whose value is an integer (1, 2, or 3).
\end{promptbox}
\begin{promptbox}{Interview Guide Coverage Judge}
You are a qualitative coding assistant. Below is a numbered list of interview guide subtopics followed by a list of interview excerpts from a single interview. Each excerpt shows the interviewer's question and the participant's response.

For each excerpt, identify every subtopic that the participant's response directly addresses. A single response may address multiple subtopics, one subtopic, or none. Base your labels on what the participant's response actually says (*\textemdash*) not on what the interviewer's question asked about. If the interviewer raises a subtopic but the participant does not engage with it, do not label it. If the response does not clearly address any subtopic, use an empty list.

Interview guide subtopics:
{subtopics_text}

Interview excerpts:
{excerpts_json}

Return a JSON object with a single key "labels" whose value is a list with one element per excerpt, in the same order as the excerpts. Each element is itself a list of the subtopic numbers (integers) that the participant's response in that excerpt directly addresses, or an empty list [] if it addresses none.
\end{promptbox}

\begin{promptbox}{Coherence Judge}
You are evaluating an interviewer's technique in a qualitative research interview.
Read the transcript below and rate the interviewer on coherence only.

Coherence: rate the extent to which the interviewer's questions logically flow, build on each other, and effectively transition between topics.
1. Mostly incoherent, most questions are disjointed and unrelated.
2. Somewhat incoherent, questions frequently jump topics without transition or conflict with prior answers.
3. Somewhat coherent, questions generally follow logically but may have minor lapses or occasional topic jumps.
4. Mostly coherent, questions consistently build on prior answers and maintain a logical flow. Topics change naturally or have clear transitions.

Transcript:
{transcript_text}

Return a JSON object with a single key "score" whose value is an integer 1-4.
\end{promptbox}
\begin{promptbox}{Adaptiveness Judge}
You are evaluating an interviewer's technique in a qualitative research interview.
Read the transcript below and rate the interviewer on adaptiveness only.

Adaptiveness: rate how well the interviewer's questions follow from and respond to the participant's answers. Weight the strongest evidence: if even one or two questions clearly engage the specific things the participant said, that is high adaptiveness, and questions that stay on the participant's topic are at least somewhat adaptive. Reserve a 1 for an interviewer who disregards the answers or follows an unrelated, fixed script.
1. Not adaptive, questions ignore the participant's answers or follow a fixed script that could be asked of anyone.
2. Somewhat adaptive, questions stay on the participant's topic but rarely engage the specifics of what they said.
3. Moderately adaptive, at least one question clearly builds on or follows up the participant's answers while others advance the guide.
4. Highly adaptive, questions repeatedly build on the participant's specific answers, probing for clarification and detail.

Transcript:
{transcript_text}

Return a JSON object with a single key "score" whose value is an integer 1-4.
\end{promptbox}
\begin{promptbox}{Leading Questions Judge}
You are evaluating an interviewer's technique in a qualitative research interview.
Read the transcript excerpt below. The interviewer's utterances are numbered [I1], [I2], ... in order. Classify each numbered interviewer utterance as either a leading question or not.

A leading question is one phrased to suggest a desired or expected answer, embed an assumption, or otherwise steer the participant toward a particular response instead of letting them answer freely. Open or neutral questions are not leading.

Transcript:
{transcript_text}

Return a JSON object with a single key "labels" whose value is a list of booleans with one element per numbered interviewer utterance, in the same order ([I1], [I2], ...). Each element is true if that utterance is a leading question and false otherwise.
\end{promptbox}
\begin{promptbox}{Support and Rapport Judge}
You are evaluating an interviewer's technique in a qualitative research interview.
Read the transcript excerpt below. The interviewer's utterances are numbered [I1], [I2], ... in order. Classify each numbered interviewer utterance as either containing a support or rapport statement or not.

A support or rapport statement is an interviewer utterance designed to make a connection with the participant, provide support, or let the participant know that the purpose of the interview is being fulfilled (e.g., affirmation, empathy, thanks, encouragement, reassurance).

Transcript:
{transcript_text}

Return a JSON object with a single key "labels" whose value is a list of booleans with one element per numbered interviewer utterance, in the same order ([I1], [I2], ...). Each element is true if that utterance contains a support or rapport statement and false otherwise.
\end{promptbox}

\begin{promptbox}{Unclear Questions Judge}
You are analyzing a qualitative interview from the participant's perspective.
Read the transcript excerpt below. The interviewer's utterances are numbered [I1], [I2], ... in order. Classify each numbered interviewer utterance as either unclear or not.

An unclear question is one that is ambiguous, confusing, double-barreled, or otherwise likely difficult for a typical participant to understand. When deciding, evaluate the question's own wording and structure (*\textemdash*) not the participant's response. A clear question that receives a vague, evasive, or brief answer is still clear; participants may give poor answers to perfectly clear questions. Mark a question unclear only if a reasonably attentive participant would find it difficult to understand or interpret on its own.

Transcript:
{transcript_text}

Return a JSON object with a single key "labels" whose value is a list of booleans with one element per numbered interviewer utterance, in the same order ([I1], [I2], ...). Each element is true if that utterance is unclear and false otherwise.
\end{promptbox}
\begin{promptbox}{Comfort Level Judge}
You are analyzing a qualitative interview from the participant's perspective.
Read the transcript below and estimate the participant's comfort level during the interview.

Transcript:
{transcript_text}

Return a JSON object with a single key "score" whose value is an integer 1-4:
  1 = Entirely uncomfortable or treated unfairly
  2 = Mostly uncomfortable, but with some exceptions
  3 = Mostly comfortable, but with some exceptions
  4 = Entirely comfortable and treated fairly
\end{promptbox}
\begin{promptbox}{Overall Experience Judge}
You are analyzing a qualitative interview from the participant's perspective.
Read the transcript below and estimate the overall quality of the interview experience from the participant's perspective.

Transcript:
{transcript_text}

Return a JSON object with a single key "score" whose value is an integer 1-5:
  1 = Poor
  2 = Below Average
  3 = Average
  4 = Good
  5 = Excellent
\end{promptbox}

\subsection{Human Validation}
\label{app:interviewreportcard-validation}

To validate the LLM judges for InterviewReportCard, we collect 1,800 expert judgments from three annotators with experience in qualitative research. Each annotator rates 50 excerpts from interviews with human participants and 50 excerpts from interviews with simulated participants from the experiments in \Sref{sec:human-experiments}. We then apply Gemini 3.1 Pro to the same excerpts, using the same information and scoring rubrics provided to the human annotators. For coherence, adaptiveness, leading questions, support and rapport, and unclear questions, we calculate Krippendorff's alpha to measure agreement among human annotators and agreement between the median human rating and the LLM judge rating. We find moderate agreement among human annotators and comparable agreement between the median human rating and the LLM judge rating (\Cref{tab:measure-validation}). 

\begin{table}[ht]
\centering
\begin{tabular}{@{}lcc@{}}
\toprule
\textbf{Measure} & \textbf{Human-Agreement} & \textbf{Human-LLM Agreement} \\
\midrule
Adaptiveness & 0.70 & 0.80 \\
Coherence & 0.65 & 0.73 \\
Leading Questions & 0.68 & 0.70 \\
Support / Rapport & 0.68 & 0.67 \\
Unclear Questions & 0.62 & 0.60 \\
\bottomrule
\end{tabular}
\caption{Agreement between human ratings (Human) and between the median human ratings and LLM judge ratings (Human-LLM) as measured with Krippendorff’s alpha}
\label{tab:measure-validation}
\end{table}

To validate the LLM judges for interview guide coverage, we compare the sets of relevant topics identified by the LLM judge with those identified by human annotators for each excerpt and find an average Jaccard similarity of $0.69$. We do not validate overall experience and comfort level against third-party judgments. Instead, we validate these measures by comparing them with participant self-reports in \Sref{sec:quantitative-results}.

\section{Experiment 1 Details}
\begin{table}[h]
\centering
\small
\begin{tabular}{@{}lcccc@{}}
\toprule
\textbf{Measure} & \textbf{Pearson $r$} & \textbf{95\% CI} & \textbf{Spearman $\rho$} & \textbf{95\% CI} \\
\midrule
Relevant Response Volume & 0.74 & [0.37, 0.91] & 0.73 & [0.32, 0.91] \\
Interview Guide Coverage & 0.95 & [0.85, 0.98] & 0.99 & [0.95, 1.00] \\
Novel Responses & 0.90 & [0.71, 0.97] & 0.93 & [0.79, 0.98] \\
Coherence & 0.95 & [0.84, 0.98] & 0.93 & [0.78, 0.98] \\
Adaptiveness & 0.92 & [0.78, 0.97] & 0.85 & [0.57, 0.95] \\
Leading Questions & 0.98 & [0.93, 0.99] & 0.91 & [0.74, 0.97] \\
Unclear Questions & 0.85 & [0.60, 0.95] & 0.83 & [0.54, 0.95] \\
Support and Rapport & 1.00 & [0.99, 1.00] & 0.94 & [0.81, 0.98] \\
Overall Experience & 0.67 & [0.25, 0.88] & 0.55 & [0.02, 0.84] \\
Comfort Level & 0.60 & [0.13, 0.85] & 0.55 & [0.01, 0.84] \\
Average Turns & 0.94 & [0.84, 0.98] & 0.91 & [0.74, 0.97] \\
Average Response Length & 0.85 & [0.61, 0.95] & 0.74 & [0.34, 0.91] \\
\bottomrule
\end{tabular}
\caption{Pearson and Spearman correlations between InterviewPlayground's simulated studies and the human studies with the same AI interviewers and interview topics.} 
\label{tab:correlation-ci-appendix}
\end{table}

\begin{table}[h]
\centering
\small
\resizebox{\linewidth}{!}{%
\begin{tabular}{lccccccc}
\toprule
\textbf{Measure} & \textbf{\shortstack{Gemini\\3.1 Pro}} & \textbf{\shortstack{GPT-5.6\\Terra}} & \textbf{\shortstack{Gemma 4\\31B}} & \textbf{\shortstack{Gemini\\3.7 Flash}} & \textbf{\shortstack{Qwen 3.5\\9B}} & \textbf{\shortstack{Qwen 3.5\\4B}} & \textbf{\shortstack{Qwen 3.5\\0.8B}} \\
\midrule
Relevant Response Volume    & 0.74 & 0.57 & 0.52 & 0.70 & 0.54 & 0.85 & 0.09 \\
Interview Guide Coverage    & 0.95 & 0.97 & 0.96 & 0.98 & 0.94 & 0.86 & 0.67 \\
Novel Responses             & 0.90 & 0.90 & 0.94 & 0.79 & 0.67 & 0.91 & 0.26 \\
Coherence                   & 0.95 & 0.95 & 0.97 & 0.96 & 0.88 & 0.77 & 0.82 \\
Adaptiveness                & 0.92 & 0.95 & 0.96 & 0.94 & 0.92 & 0.83 & 0.80 \\
Leading Questions           & 0.98 & 0.97 & 0.96 & 0.97 & 0.97 & 0.95 & 0.93 \\
Unclear Questions           & 0.85 & 0.94 & 0.90 & 0.91 & 0.70 & 0.94 & 0.58 \\
Support and Rapport         & 1.00 & 0.98 & 0.98 & 0.99 & 0.97 & 0.93 & 0.93 \\
Overall Experience          & 0.67 & 0.63 & 0.56 & 0.73 & 0.67 & 0.86 & 0.39 \\
Comfort Level               & 0.60 & 0.38 & 0.51 & 0.66 & 0.37 & 0.96 & 0.48 \\
Average Turns               & 0.94 & 0.91 & 0.92 & 0.91 & 0.84 & 0.41 & 0.76 \\
Average Response Length     & 0.85 & 0.71 & 0.66 & 0.77 & 0.80 & 0.73 & 0.79 \\
\midrule
\textbf{Average Correlation}            & \textbf{0.86} & \textbf{0.82} & \textbf{0.82} & \textbf{0.86} & \textbf{0.77} & \textbf{0.83} & \textbf{0.62} \\
\bottomrule
\end{tabular}%
}
\caption{Pearson correlation between the results of InterviewPlayground's simulated studies and the human studies across the 12 measures using seven different LLMs as the simulator model.}
\label{tab:robustness}
\end{table}

\end{document}